\documentclass{sig-alternate-2013}
\usepackage[ruled]{algorithm2e}
\usepackage{subfigure}
\usepackage{multirow}
\usepackage{color, verbatim,caption}
\usepackage{hyperref}
\newtheorem{definition}{Definition}

\newfont{\mycrnotice}{ptmr8t at 7pt}
\newfont{\myconfname}{ptmri8t at 7pt}
\let\crnotice\mycrnotice%
\let\confname\myconfname%

\permission{Permission to make digital or hard copies of all or part of this work for personal or classroom use is granted without fee provided that copies are not made or distributed for profit or commercial advantage and that copies bear this notice and the full citation on the first page. Copyrights for components of this work owned by others than ACM must be honored. Abstracting with credit is permitted. To copy otherwise, or republish, to post on servers or to redistribute to lists, requires prior specific permission and/or a fee. Request permissions from Permissions@acm.org.}
\conferenceinfo{KDD'15,}{August 10-13, 2015, Sydney, NSW, Australia.} 
\copyrightetc{\copyright~2015 ACM. ISBN \the\acmcopyr}
\crdata{978-1-4503-3664-2/15/08\ ...\$15.00.\\
	DOI:  http://dx.doi.org/10.1145/2783258.2783307}

\begin{document}

\title{PTE: Predictive Text Embedding through Large-scale Heterogeneous Text Networks}

\numberofauthors{3}
\author{	
\alignauthor
Jian Tang\\
       \affaddr{Microsoft Research Asia}\\
       \email{ jiatang@microsoft.com}
\alignauthor
Meng Qu\thanks{This work was done when the second author was an intern at Microsoft Research Asia.}\\
       \affaddr{Peking University}\\
       \email{mnqu@pku.edu.cn}
\alignauthor
Qiaozhu Mei\\
       \affaddr{University of Michigan}\\
       \email{qmei@umich.edu}
}

\maketitle
\begin{abstract}

Unsupervised text embedding methods, such as Skip-gram and Paragraph Vector, have been attracting increasing attention due to their simplicity, scalability, and effectiveness. However, comparing to sophisticated deep learning architectures such as convolutional neural networks, these methods usually yield inferior results when applied to particular machine learning tasks. One possible reason is that these text embedding methods learn the representation of text in a fully unsupervised way, without leveraging the labeled information available for the task. 
Although the low dimensional representations learned are applicable to many different tasks, they are not particularly tuned for any task.  
In this paper, we fill this gap by proposing a semi-supervised representation learning method for text data, which we call the \textit{predictive text embedding} (PTE). Predictive text embedding utilizes both labeled and unlabeled data to learn the embedding of text. The labeled information and different levels of word co-occurrence information are first represented as a large-scale heterogeneous text network, which is then embedded into a low dimensional space  through a principled and efficient algorithm. This low dimensional embedding not only preserves the semantic closeness of words and documents, but also has a strong predictive power for the particular task. Compared to recent supervised approaches based on convolutional neural networks, predictive text embedding is comparable or more effective, much more efficient, and has fewer parameters to tune. 


\end{abstract}
\vskip 5pt

\section{Introduction}
\label{sec::intro}
Learning a meaningful and effective representation of text, e.g., for words and documents, is a critical prerequisite for many machine learning tasks such as text classification, clustering and retrieval. Traditionally, every word is represented independently to each other, and each document is represented as a ``bag-of-words''. However, both representations suffer from problems such as data sparsity, polysemy, and synonymy,  as the semantic relatedness between different words are commonly ignored. 

\begin{figure*}[htdb!]
	\centering
	\includegraphics[width=1\textwidth]{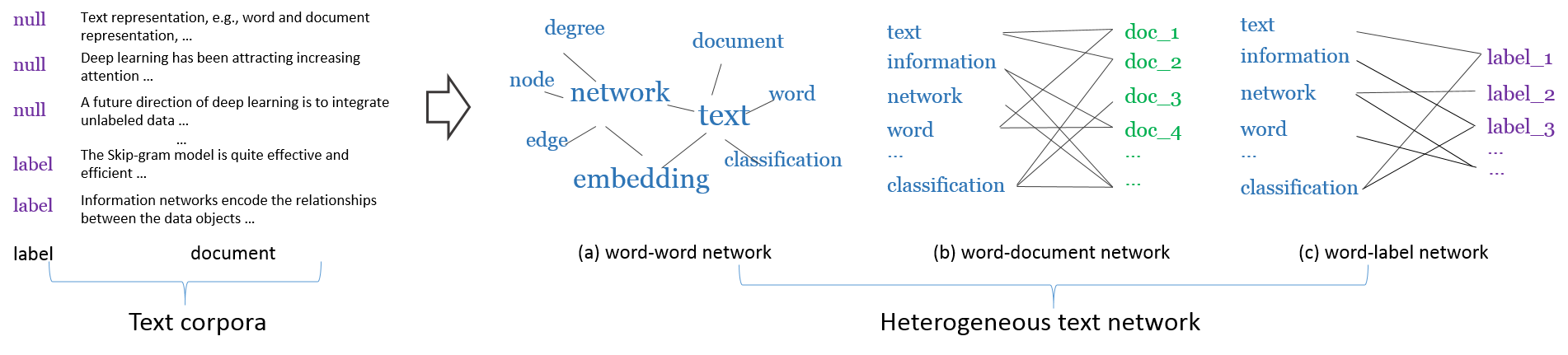}
	\caption{Illustration of converting a partially labeled text corpora to a heterogeneous text network. The word-word co-occurrence network and word-document network encode the unsupervised information, capturing the local context-level and document-level word co-occurrences respectively; the word-label network encodes the supervised information, capturing the class-level word co-occurrences.}
	\label{fig::text2netwok}
\end{figure*}

Distributed representations of words and documents~\cite{mikolov2013distributed,le2014distributed} effectively address this problem through representing words and documents in low-dimensional spaces, in which similar words and documents are embedded closely to each other. The essential idea of these approaches comes from the distributional hypothesis that ``you shall know a word by the company it keeps'' (Firth, J.R. 1957:11)~\cite{Firth1957}. Mikilov et al. proposed a simple and elegant word embedding model called the Skip-gram~\cite{mikolov2013distributed}, which uses the embedding of the target word to predict the embedding of each individual context word in a local window. Le and Mikolov further extended this idea and proposed the Paragraph Vectors~\cite{le2014distributed} in order to embed arbitrary pieces of text, e.g., sentences and documents. The basic idea is to use the embeddings of sentences/documents to predict the embeddings of words in the sentences/documents. Comparing to other classical approaches that also utilize the distributional similarity of word context, such as the Brown clustering or nearest neighbors, these text embedding approaches have been proved to be quite efficient, scaling up to millions of documents on a single machine~\cite{mikolov2013distributed}. 

Because of the unsupervised learning process, the representations learned through these text embedding models are general enough and can be applied to a variety of tasks such as classification, clustering and ranking. However, when compared end-to-end with sophisticated deep learning approaches such as the convolutional neural networks (CNNs) ~\cite{blunsom2014convolutional,kim2014convolutional}, 
the performance of text embeddings usually falls short on specific tasks \cite{weston2014tagspace}. 
This is perhaps not surprising as the deep neural networks fully leverage labeled information that is available for a task when they learn the representations of the data. Most text embedding methods are not able to consider labeled information when learning the representations; the labels only kick in later when a classifier is trained using the representations as features. In other words, unsupervised text embeddings are generalizable for different tasks but have a weaker predictive power for a particular task. 

Despite this deficiency, there are still considerable advantages of text embedding approaches comparing to deep neural networks. First, the training of deep neural networks, especially convolutional neural networks is computational intensive, which usually requires multiple GPUs or clusters of CPUs when processing a large amount of data; second, convolutional neural networks usually assume the availability of a large amount of labeled examples which is unrealistic in many tasks. The easily obtainable unlabeled data are usually used through an indirect way of pre-training; third, the training of CNNs requires exhaustive tuning of many parameters, which is very time consuming even for experts and infeasible for non-experts. On the other hand, text embedding methods like Skip-gram are much more efficient, are much easier to tune, and naturally accommodate unlabeled data. 

In this paper, we fill this gap by proposing the predictive text embedding (PTE), which 
adapts the advantages of unsupervised text embeddings but naturally utilizes labeled information in representation learning. With predictive text embedding, an effective low dimensional representation is learned jointly from limited labeled examples and a large amount of unlabeled examples. Comparing to unsupervised embeddings, this representation is optimized for particular tasks like what convolutional neural networks do (i.e., the representation has strong predictive power for the particular classification task). 

The proposed method naturally extends our previous work of unsupervised information network embedding \cite{tang2015line} and first learns a low dimensional embedding for words through a heterogeneous text network. The network encodes different levels of co-occurrence information between words and words, words and documents, and words and labels. The network is embedded into a low dimensional vector space that preserves the second-order proximity~\cite{tang2015line} between the vertices in the network. The representation of an arbitrary piece of text (e.g., a sentence or a document) can be simply inferred as the average of the word representations, which turns out to be quite effective. 
The whole optimization process remains very efficient, which scales up to millions of documents and billions of tokens on a single machine. 

We conduct extensive experiments with real-world text corpora, including both long and short documents. Experimental results show that the predictive text embeddings significantly outperform the state-of-the-art unsupervised embeddings in various text classification tasks. Compared end-to-end with convolutional neural networks for text classification~\cite{kim2014convolutional}, predictive text embedding outperforms on long documents and generates comparable results on short documents. PTE enjoys various advantages over convolutional neural networks as it is much more efficient, accommodates large-scale unlabeled data effectively, and is less sensitive to model parameters. 
We believe our exploration points to a direction of learning text embeddings that could compete head-to-head with deep neural networks in particular tasks. 

To summarize, we make the following contributions:
\begin{itemize}
	\item We propose to learn predictive text embeddings in a semi-supervised manner. Unlabeled data and labeled information are integrated into a heterogeneous text network which incorporates different levels of co-occurrence information in text.
	\item We propose an efficient algorithm ``PTE'', which learns a distributed representation of  text through embedding the heterogeneous text network into a low dimensional space. The algorithm is very efficient and has few parameters to tune. 
	\item We conduct extensive experiments using various real-world data sets and compare predictive text embedding end-to-end with both unsupervised text embeddings and convolutional neural networks. 
\end{itemize}

The rest of this paper is organized as follows. We first introduce the related work in Section~\ref{sec::related}. Section~\ref{sec::definition} formally defines the problem of predictive text embedding through heterogeneous text networks. Section~\ref{sec::model} introduces the proposed algorithm in details. Section~\ref{sec::experiment} presents the results of empirical experiments. We conclude in Section~\ref{sec::conclusion}.

\section{Related Work}
\label{sec::related}
Our work is mainly related to distributed text representation learning and information network embedding. 

\subsection{Distributed Text Embedding}
Distributed representation of text has proved to be quite effective in many natural language processing tasks such as word analogy~\cite{mikolov2013distributed}, POS tagging~\cite{collobert2011natural}, parsing~\cite{collobert2011natural}, language modeling~\cite{mikolov2010recurrent}, and sentiment analysis~\cite{mesnil2014ensemble,le2014distributed,blunsom2014convolutional,kim2014convolutional}. Existing approaches can be generally classified into two categories: unsupervised and supervised. Recent developed unsupervised approaches normally learn the embeddings of words and/or documents by utilizing word co-occurrences in the local context (e.g., Skip-gram~\cite{mikolov2013distributed}) or at document level (e.g., paragraph vectors~\cite{le2014distributed}). These approaches are quite efficient, scaling up to millions of documents. The supervised approaches~\cite{blunsom2014convolutional,kim2014convolutional,shen2014latent,collobert2011natural} are usually based on deep neural network architectures, such as recursive neural tensor networks (RNTNs)~\cite{socher2013recursive} or convolutional neural networks (CNNs)~\cite{lecun1995convolutional}. 
In RNTNs, each word is embedded into a low dimensional vector, and the embeddings of the phrases are recursively learned by applying the same tensor-based composition function over the sub-phrases or words in a parse tree. In CNNs~\cite{blunsom2014convolutional}, each word is also represented with a vector, and the same convolutional kernel is applied over the context windows in different positions of the sentences, followed by a max-pooling and fully connected layer. 

The major difference between these two categories of approaches is how they utilize labeled and unlabeled information in the representation learning phase. The unsupervised methods do not include labeled information when learning the representations and only use the labels to train the classifier after the data is transformed into the learned representation. RNTNs and CNNs incorporate the labels directly into representation learning, so the learned representations are particularly tuned for the classification task. To incorporate unlabeled examples, however, these neural nets usually have to use an indirect approach such as to pretrain the word embeddings with unsupervised approaches. Comparing to these two lines of work, PTE learns the text vectors in a semi-supervised way - the representation learning algorithm directly utilizes both labeled information and large-scale unlabeled data. 


Another piece of work similar to predictive word embedding is~\cite{maas2011learning}, which learns word vectors that are particularly tuned for sentiment analysis. However, their approach does not scale to millions of documents and does not generalize to other classification tasks.

\subsection{Information Network Embedding}
Our work is also related to the problem of network/graph embedding as the word representations of PTE are learned through a heterogeneous text network. Embedding networks/graphs into low dimensional spaces is very useful in a variety of applications, e.g., node classification~\cite{bhagat2011node} and link prediction~\cite{liben2007link}. Classical graph embedding algorithms such as MDS~\cite{joseph1978multidimensional}, IsoMap~\cite{tenenbaum2000global} and Laplacian eigenmap~\cite{belkin2001laplacian} are not applicable for embedding large-scale networks that contain millions of vertices and billions of edges. There are some recent work attempting to embed very large real-world networks. Perozzi et al. ~\cite{perozzi2014deepwalk} proposed a network embedding model called the ``DeepWalk,'' which uses truncated random walks on the networks and is only applicable for networks with binary edges. Our previous work proposed a novel large-scale network embedding model called the ``LINE,'' which is suitable for arbitrary types of information networks: undirected or directed, binary or weighted~\cite{tang2015line}. The LINE model optimizes an objective function which aims to preserve both the local and global network structures. Both Deepwalk and LINE are unsupervised and only handle homogeneous networks. The network embedding algorithm used by PTE extends the LINE to deal with heterogeneous networks, in which multiple types of vertices (including the class labels) and edges exist.

\section{Problem Definition} 
\label{sec::definition}
Let us begin with formally defining the problem of predictive text embedding through heterogeneous text networks. Comparing to unsupervised text embedding approaches including Skip-gram and Paragraph Vectors that learn general semantic representations of text, 
our goal is to learn a representation of text that is optimized for a given text classification task. In other words, we anticipate the text embedding to have a strong predictive power of the performance of the given task. 
The basic idea is to incorporate both the labeled and unlabeled information when learning the text embeddings. To achieve this, it is desirable to first have an unified representation to encode both types of information. In this paper, we propose different types of networks to achieve this, including word-word co-occurrence networks, word-document networks, and word-label networks. 

\begin{definition}
	\label{def::ww-net}
	\textbf{(Word-Word Network)}
	\textsl{\textbf{Word-word} co-occurrence network, denoted as $G_{ww}=(\mathcal{V},E_{ww})$, captures the word co-occurrence information in local contexts of the unlabeled data. $\mathcal{V}$ is a vocabulary of words and $E_{ww}$ is the set of edges between words. The weight $w_{ij}$ of the edge between word $v_i$ and $v_j$ is defined as the number of times that the two words co-occur in the context windows of a given window size.}
\end{definition}

The word-word network captures the word co-occurrences in local contexts, which is the essential information used by existing word embedding approaches such as Skip-gram. Beyond the local contexts, word co-occurrence at the document level is also widely explored in classical text representations such as statistical topic models, e.g., the latent Dirichlet allocation~\cite{blei2003latent}. To capture the document-level word co-occurrences, we introduce another network, word-document network, defined as below:

\begin{definition}
	\label{def::wd-net}
	\textbf{(Word-Document Network)}
	\textsl{\textbf{Word-\\ document} network, denoted as $G_{wd}=(\mathcal{V} \cup \mathcal{D},E_{wd})$, is a bipartite network where $\mathcal{D}$ is a set of documents and $\mathcal{V}$ is a set of words. $E_{wd}$ is the set of edges between words and documents. The weight $w_{ij}$ between word $v_i$ and document $d_j$ is simply defined as the number of times $v_i$ appears in document $d_j$.} 
\end{definition}

The word-word and word-document networks encode the unlabeled information in large-scale corpora, capturing word co-occurrences at both the local context level and the document level. To encode the labeled information, we introduce the word-label network, which captures word co-occurrences at category-level .

\begin{definition}
	\label{def::wlnet}
	\textbf{(Word-Label Network)}
	\textsl{\textbf{Word-label} network, denoted as $G_{wl}=(\mathcal{V} \cup \mathcal{L},E_{wl})$, is a bipartite network that captures category-level word co-occurrences. $\mathcal{L}$ is a set of class labels and $\mathcal{V}$ a set of words. $E_{wl}$ is a set of edges between words and classes. The weight $w_{ij}$ of the edge between word $v_i$ and class $c_j$ is defined as: $w_{ij}=\sum_{(d:l_d=j)} n_{di}$, where $n_{di}$ is the term frequency of word $v_i$ in document $d$, and $l_d$ is the class label of document $d$.}
\end{definition}

The three types of networks above can be further integrated into one heterogeneous text network.

\begin{definition}
	\label{def::wdlnet}
	\textbf{(Heterogeneous Text Network)}
	\textsl{The \textbf{heterogeneous text network} is the combination of word-word, word-document, and word-label networks constructed from both unlabeled and labeled text data. It captures different levels of word co-occurrences and contains both labeled and unlabeled information. }
\end{definition}

Note that the definition of a heterogeneous text network can be generalized to integrate other types of networks such as word-sentence, word-paragraph, and document-label networks. In this work we are using the three types of networks (word-word, word-document, and word-label) as an illustrative example. We particularly focus on word networks in order to first represent words into low dimensional spaces. The representation of other text units (e.g., sentences or paragraphs) can be then computed through aggregating the word representations. 

Finally, we formally define the problem of predictive text embedding as follows:

\begin{definition}
	\label{def::wdlnet}
	\textbf{(Predictive Text Embedding)}
	\textsl{Given a large collection of text data with unlabeled and labeled information, the problem of \textbf{predictive text embedding} aims to learn low dimensional representations of words by embedding the heterogeneous text network constructed from the collection into a low dimensional vector space. }
\end{definition}


\section{Predictive Text Embedding}
\label{sec::model}
In this section, we introduce the proposed method that learns predictive text embedding through heterogeneous text networks. Our method first learns vector representations of words by embedding the heterogeneous text networks constructed from free text into a low dimensional space, and then infer text embeddings based on the learned word vectors. As the heterogeneous text network is composed of three bipartite networks, we first introduce an approach for embedding individual bipartite networks.

\subsection{Bipartite Network Embedding}
In our previous work, we introduced the LINE model to learn the embedding of large-scale information networks~\cite{tang2015line}. LINE is mainly designed for homogeneous networks, i.e., networks with the same types of nodes. LINE cannot be directly applied to heterogeneous networks as the weights on different types of edges are not comparable. Here, we first adapt the LINE model for embedding bipartite networks. The essential idea is to make use of the second-order proximity~\cite{tang2015line} between vertices, which assumes vertices with similar neighbors are similar to each other and thus should be represented closely in a low dimensional space.  

Given a bipartite network $G=(\mathcal{V}_A \cup \mathcal{V}_B, E)$, where $\mathcal{V}_A$ and $\mathcal{V}_B$ are two disjoint sets of vertices of different types, and $E$ is the set of edges between them. We first define the conditional probability of vertex $v_i$ in set $\mathcal{V}_A$ generated by vertex $v_j$ in set $\mathcal{V}_B$ as:

\begin{equation}
	\label{eqn::2nd_prob}
	p(v_i|v_j)=\frac{\exp(\vec{u}_i^T\cdot \vec{u}_j)}{\sum_{i'\in A} \exp({\vec{u}_{i'}}^T \cdot \vec{u}_j)},
\end{equation}
where $\vec{u}_i$ is the embedding vector of vertex $v_i$ in $\mathcal{V}_A$, and $\vec{u}_j$ is the embedding vector of vertex $v_j$ in $\mathcal{V}_B$. For each vertex $v_j$ in $\mathcal{V}_B$, Eq~\eqref{eqn::2nd_prob} defines a conditional distribution $p(\cdot|v_j)$ over all the vertices in the set $\mathcal{V}_A$; for each pair of vertices $v_j, v_{j'}$, the second-order proximity can actually be determined by their conditional distributions $p(\cdot|v_j), p(\cdot|v_{j'})$. To preserve the second-order proximity, we can make the conditional distribution $p(\cdot|v_j)$ be close to its empirical distribution $\hat{p}(\cdot|v_j)$, which can be achieved by minimizing the following objective function:

\begin{equation}
	\label{eqn::line_obj}
	O=\sum_{j\in B} \lambda_j d(\hat{p}(\cdot|v_j), p(\cdot|v_j)),
\end{equation}
where $d(\cdot,\cdot)$ is the KL-divergence between two distributions, $\lambda_j$ is the importance of vertex $v_j$ in the network, which can be set as the degree $deg_j=\sum_i w_{ij}$, and the empirical distribution can be defined as $\hat{p}(v_i|v_j)=\frac{w_{ij}}{deg_j}$. Omitting some constants, the objective function~\eqref{eqn::line_obj} can be calculated as:

\begin{equation}
	\label{eqn::line_obj2}
	O=-\sum_{(i,j)\in E} w_{ij}\log p(v_j|v_i).
\end{equation}

The objective~\eqref{eqn::line_obj2} can be optimized with stochastic gradient descent using the techniques of edge sampling~\cite{tang2015line} and negative sampling~\cite{mikolov2013distributed}. In each step, a binary edge $e=(i,j)$ is sampled with the probability proportional to its weight $w_{ij}$, and meanwhile multiple negative edges $(i,j)$ are sampled from a noise distribution $p_n(j)$. The sampling procedures address significant deficiency of stochastic gradient descent in learning network embeddings. For the detailed optimization process, readers can refer to~\cite{tang2015line}.

The embeddings of the word-word, word-document, and word-label network can all be learned by the above model. Note that the word-word network is essentially a bipartite-network by treating each undirected edge as two directed edges, and then $\mathcal{V}_A$ is defined as the set of the source nodes, $\mathcal{V}_B$ as the set of target nodes. Therefore, we can define the conditional probabilities $p(v_i|v_j)$, $p(v_i|d_j)$ and $p(v_i|l_j)$ according to equation~\eqref{eqn::2nd_prob}, and then learn the embeddings by optimizing objective function~\eqref{eqn::line_obj2}. Next, we introduce our approach of embedding the heterogeneous text network. 

\subsection{Heterogeneous Text Network Embedding}
The heterogeneous text network is composed of three bipartite networks: word-word, word-document and word-label networks, where the word vertices are shared across the three networks. To learn the embeddings of the heterogeneous text network, an intuitive approach is to collectively embed the three bipartite networks, which can be achieved by minimizing the following objective function:

\begin{equation}
	\label{eqn::obj_lhine}
	O_{pte}= O_{ww}+O_{wd}+O_{wl},
\end{equation}
where
\begin{equation}
	O_{ww}=-\sum_{(i,j)\in E_{ww}} w_{ij} \log p(v_i|v_j)
\end{equation}
\begin{equation}
	O_{wd}=-\sum_{(i,j)\in E_{wd}} w_{ij}\log p(v_i|d_j)
\end{equation}
\begin{equation}
	O_{wl}=-\sum_{(i,j)\in E_{wl}} w_{ij}\log p(v_i|l_j)
\end{equation}

The objective function~\eqref{eqn::obj_lhine} can be optimized in different ways, depending on how the labeled information, i.e., the word-label network, is used. One solution is to train the model with the unlabeled data (the word-word and word-document networks) and the labeled data simultaneously. We call this approach \emph{joint training}. An alternative solution is to learn the embeddings with unlabeled data first, and then fine-tune the embeddings with the word-label network. This is inspired by the idea of \textit{pre-training and fine-tuning} in the literature of deep learning~\cite{bengio2013representation}.

In joint training, all three types of networks are used together. A straightforward solution to optimize the objective~\eqref{eqn::obj_lhine} is to merge the all the edges in the three sets $E_{ww}, E_{wd}, E_{wl}$ and then deploy edge sampling~\cite{tang2015line}, which samples an edge for model updating in each step, with the sampling probability proportional to its weight. However, when the network is heterogeneous, the weights of the edges between different types of vertices are not comparable to each other. A more reasonable solution is to alternatively sample from the three sets of edges.  We summarize the detailed training algorithm in Alg.~\ref{algo::joint}.

\begin{algorithm}[!htdb]
	\scriptsize
	\KwData{$G_{ww},G_{wd},G_{wl}$, number of samples $T$, number of negative samples $K$.}
	\KwResult{word embeddings $\vec{w}$.}
	\While{ iter $\leq$ $T$}{
		\begin{itemize}
			\item	sample an edge from $E_{ww}$ and draw $K$ negative edges, \\ and update the word embeddings\;
			\item	sample an edge from $E_{wd}$ and draw $K$ negative edges, \\ and update the word and document embeddings\;
			\item 	sample an edge from $E_{wl}$ and draw $K$ negative edges, \\ and update the word and label embeddings\;
		\end{itemize}
	}
	\caption{Joint training.}
	\label{algo::joint}
\end{algorithm}
\vskip -2em
\begin{algorithm}[!htdb]
	\scriptsize
	\KwData{$G_{ww},G_{wd},G_{wl}$, number of samples $T$, number of negative samples $K$.}
	\KwResult{word embeddings $\vec{w}$.}
	\While{ iter $\leq$ $T$}{
		\begin{itemize}
			\item sample an edge from $E_{ww}$ and draw $K$ negative edges, \\ and update the word embeddings\;
			\item sample an edge from $E_{wd}$ and draw $K$ negative edges, \\ and update the word and document embeddings\;
		\end{itemize}
	}
	\While{ iter $\leq$ $T$}{
		\begin{itemize}
			\item sample an edge from $E_{wl}$ and draw $K$ negative edges, \\ and update the word and label embeddings\;
		\end{itemize}
	}	
	\caption{Pre-training + Fine-tuning.}
	\label{algo::separate}
\end{algorithm}

Similarly, we summarize the training process of pre-training and fine-tuning in Alg.~\ref{algo::separate}.

\subsection{Text Embedding}

The heterogeneous text network encodes word co-occurrences at different levels, extracted from both unlabeled data and labeled information for a specific classification task. Therefore, the word representations learned by embedding the heterogeneous text network are not only more robust but also optimized for that task.
Once the word vectors are learned, the representation of an arbitrary piece of text can be obtained by simply averaging the vectors of the words in that piece of text. That is, the vector representation of a piece of text $d=w_1w_2\cdots,w_n$ can be computed as  
\begin{equation}
	\vec{d}=\frac{1}{n}\sum_{i=1}^n \vec{u}_i,
\end{equation} 
where $\vec{u}_i$ is the embedding of word $w_i$.

In fact, the average of the word embeddings is the solution to minimizing the following objective function:

\begin{equation}
	\label{eqn::infer_obj}
	O=\sum_{i=1}^{n} l(\vec{u}_i, \vec{d}),
\end{equation}
where the loss function $l(\cdot,\cdot)$ between the word embedding $\vec{u}_i$ and text embedding $\vec{d}$ is specified as the Euclidean distance. Related is the inference process of paragraph vectors~\cite{le2014distributed}, which minimizes the same objective but with a different loss function $l(\vec{u}_i, \vec{d})=-\frac{1}{1+\exp(-\vec{u}_i^T\vec{d})}$. It however does not lead to a close form solution and has to be optimized by gradient descent algorithm.

\section{Experiments}
\label{sec::experiment}
In this section, we move forward to evaluate the effectiveness of the proposed PTE algorithm for predictive text embedding. A variety of text classification tasks and data sets are selected for this purpose. The experiments are set up as the following.

\subsection{Experiment Setup}
\subsubsection*{Data Sets}
We select two types of text corpora, which consist of either long or short documents. 

\noindent \textbf{Long Document Corpora:} (1) \textsc{20ng,} the widely used text classification data set 20newsgroup\footnote{\scriptsize Available at \url{http://qwone.com/~jason/20Newsgroups/}}, containing 20 categories; (2)\textsc{Wiki,} a snapshot of Wikipedia corpus in April 2010 containing around two million English articles. Only common words appeared in the vocabulary of \emph{wiki2010}~\cite{shaoul2010westbury} are kept. We choose seven diverse categories for the classification task, including ``Arts,'' ``History,'' ``Human,'' ``Mathematics,'' ``Nature," ``Technology,'' and ``Sports'' from DBpedia ontology\footnote{\scriptsize Available at \url{http://downloads.dbpedia.org/3.9/en/article_categories_en.nq.bz2.}}. For each category, we randomly select 9,000 articles as labeled documents for training; (3) \textsc{Imdb,} a data set for sentiment classification from~\cite{maas2011learning}\footnote{\scriptsize Available at~\url{http://ai.stanford.edu/~amaas/data/sentiment/}}. To avoid the distribution bias between the training and test data sets, we randomly shuffle the training and test data sets; (4) \textsc{RCV1,} a large benchmark corpus for text classification~\cite{lewis2004rcv1}\footnote{\scriptsize Available at \url{http://www.ai.mit.edu/projects/jmlr/papers/volume5/lewis04a/lyrl2004_rcv1v2_README.htm}}. Four subsets including \textsc{Corporate}, \textsc{Economics}, \textsc{Government} and \textsc{Market} are extracted from the corpus. In \textsc{RCV1} data sets, all the documents have already been represented as ``bag-of-words,'' and orders between words are lost. 

\noindent \textbf{Short Document Corpora:} (1) \textsc{Dblp,} which contains titles of papers from the computer science bibliography\footnote{\scriptsize Available at \url{http://arnetminer.org/billboard/citation}}. We choose six diverse research fields for classification including ``database,'' ``artificial intelligence,'' ``hardware,'' ``system,'' ``programming languages,'' and ``theory.'' For each field, we select representative conferences and collect the papers published in the selected conferences as the labeled documents; (2) \textsc{Mr,} a movie review data set, in which each review only contains one sentence~\cite{pang2005seeing}\footnote{\scriptsize Available at \url{http://www.cs.cornell.edu/people/pabo/movie-review-data/}}; (3) \textsc{Twitter,} a corpus of Tweets for sentiment classification\footnote{\scriptsize Available at \url{http://thinknook.com/ twitter-sentiment-analysis-training-corpus-dataset-2012-09-22/}}, from which we randomly sampled 1,200,000 Tweets and split them into training and testing sets.

No further text normalization such as removing stop words or stemming is done on top of the original data. We summarize the detailed statistics of these data sets in Table~\ref{tab::dataset-statistics}.

\begin{table*}[bht!]
	\caption{Statistics of the Data Sets}
	\label{tab::dataset-statistics}
	\begin{center}
		\scalebox{0.8}{
			\begin{tabular}{c|c|c|c|c|c|c|c|c|c|c} \hline
				&\multicolumn{7}{|c|}{\textbf{Long Documents}}&\multicolumn{3}{|c}{\textbf{Short Documents}} \\ \hline
				Name&\textsc{20ng}&\textsc{Wiki}&\textsc{IMDB}&\textsc{Corporate}&\textsc{Economics}&\textsc{Government}&\textsc{Market}&\textsc{Dblp}&\textsc{Mr}&\textsc{Twitter}\\ \hline
				Train&11,314  &1,911,617*  &25,000 & 245,650  &77,242  &138,990  &132,040 &61,479  &7,108  &800,000\\ \hline 
				Test  &7,532  &21,000  &25,000 & 122,827  &38,623  &69,496  &66,020 &20,000  &3,554  &400,000\\ \hline 				
				|V|   &89,039  &913,881  &71,381 & 141,740  &65,254  &139,960  &64,049 &22,270  &17,376  &405,994\\ \hline 
				Doc. length&305.77&672.56 &231.65 & 102.23  &145.10 &169.07  &119.83 &9.51  &22.02  &14.36\\ \hline 
				\#classes& 20 &7 &2 &18 & 10  &23  &4 &6  &2  &2\\ \hline 				
			\end{tabular}
		}
		\\	\scriptsize *In the \textsc{Wiki} data set, only 42,000 documents are labeled.			
	\end{center}
\end{table*}

\subsubsection*{Compared Algorithms}
We compare the PTE algorithm with other representation learning algorithms for text data, including the classical ``bag-of-words'' representation and the state-of-the-art approaches to unsupervised and supervised text embedding.

\begin{itemize}
	\item BOW: the classical ``bag-of-words'' representation. Each document is represented with a $|V|$-dimensional vector, in which the weight of each dimension is calculated with the TFIDF weighting~\cite{salton1988term}.
	\item Skip-gram: the state-of-the-art word embedding model proposed by Mikolov et al.~\cite{mikolov2013distributed}. For the document embedding, we simply take the average of the word embeddings as explained in Section 4.3.
	\item PVDBOW: the distributed bag-of-words version of paragraph vector model proposed by Le and Mikolv~\cite{le2014distributed}, in which the orders of words in a document are ignored. 
	\item PVDM: the distributed memory version of paragraph vector which considers the order of the words~\cite{le2014distributed}.
	\item LINE: the large-scale information network embedding model proposed by Tang et al.~\cite{tang2015line}. We use the LINE model to learn unsupervised embeddings with the word-word network, word-document network or the combination of the two networks.
	\item CNN: the supervised text embedding approach based on a convolutional neural network~\cite{kim2014convolutional}. Though CNN is proposed for modeling sentences, we adapt it for general word sequences including long documents. Although CNN typically works with fully labeled documents, it can also utilize unlabeled data by pre-training the model with unsupervised word embeddings, which is marked as CNN(pretrain).
	\item PTE: our proposed approach for learning predictive text embedding. There are different variants of PTE that use different combinations of the word-word, word-document and word-label networks. We denote PTE($G_{wl}$) for the version that uses the word-label network only; PTE(pretrain) learns an unsupervised embedding with the word-word and word-document networks, and then fine-tune the word embeddings with the word-label network; PTE(joint) jointly trains the heterogeneous text network composed of all the three networks.	 
\end{itemize}

\subsubsection*{Classification and Parameter Settings}
Once the vector representations of documents are constructed or learned, we apply the same classification process using the same training data set. In particular, all the documents in the training set are used in both the representation learning phase and the classifier learning phase. The class labels of these documents are not used in the representation learning phase if an unsupervised embedding method is used; they only kick in at the classifier learning phase. The class label are used in both the representation learning phase and the classifier learning phase if a predictive embedding method is used. The test data is held-out from both phases. In the classification phase, we use the one-vs-rest logistic regression model in the LibLinear package\footnote{\url{http://www.csie.ntu.edu.tw/~cjlin/liblinear/}}. 
The classification performance is measured with the micro-F1 and macro-F1 metrics. For Skip-gram, PVDBOW, PVDM and PTE, the mini-batch size of the stochastic gradient descent is set as 1; the learning rate is set as $\rho_t=\rho_0(1-t/T)$, in which $T$ is the total number of mini-batches or edge samples and $\rho_0=0.025$; the number of negative samples is set as 5; the window size is set as 5 in Skip-gram, PVDBOW, PVDM and when constructing the word-word co-occurrence network. We use the structure of the CNN in~\cite{kim2014convolutional}, which uses one convolution layer, followed by a max-pooling layer and a fully-connected layer. Following ~\cite{kim2014convolutional}, we set the window size in the convolution layer as 3 and the number of feature maps as 100. For CNN, $1\%$ of the training data set is randomly selected as the validation data for early stopping. The dimensionality of word vectors is set as 100 by default for all the embedding models. 

Note that for the PTE models, the parameters are all set as above by default on different data sets. The only parameter that needs to be tuned is the number of samples $T$ in edge sampling, which can be safely set to be large.

\subsection{Quantitative Results}
\subsubsection{Performance on Long Documents}

\begin{table*} [!htdb]
	\caption{Results of text classification on \textbf{long} documents. }
	\label{fig::results-long1}
	\begin{center}
		\scalebox{0.9}{
			\begin{tabular}{l|c|c|c|c|c|c|c}\hline
				& & \multicolumn{2}{c|}{20NG} & \multicolumn{2}{|c|}{Wikipedia}& \multicolumn{2}{|c}{IMDB}\\\hline
				Type&Algorithm&  Micro-F1& Macro-F1 &  Micro-F1& Macro-F1 &  Micro-F1& Macro-F1 \\ \hline \hline
				\scriptsize Word &BOW&80.88 &79.30 &79.95 &80.03 &86.54  &86.54 \\ \hline
				\multirow{6}{0.6in}{\scriptsize Unsupervised \\ Embedding}		&Skip-gram&70.62 &68.99 &75.80 &75.77 & 85.34 &85.34  \\ \cline{2-8}								
				&PVDBOW&75.13 &73.48 &76.68 &76.75 &86.76  &86.76 \\ \cline{2-8}		
				&PVDM&61.03 &56.46 &72.96 &72.76 & 82.33 & 82.33\\ \cline{2-8}						
				&LINE($G_{ww}$)&72.78 &70.95 &77.72 &77.72 & 86.16 &86.16 \\ \cline{2-8}
				&LINE($G_{wd}$)&79.73 &78.40 &80.14 & 80.13 &89.14  & 89.14\\ \cline{2-8}
				&LINE($G_{ww}+G_{wd}$)&78.74 &77.39 &79.91 &79.94 & 89.07 &89.07 \\ \hline
				\multirow{7}{0.6in}{\scriptsize Predictive \\ Embedding}			&CNN&78.85&78.29&79.72 & 79.77 & 86.15 &86.15 \\ \cline{2-8}
				&CNN(pretrain)&80.15 &79.43 & 79.25 &79.32  &89.00  &89.00 \\ \cline{2-8}						
				&PTE($G_{wl}$)&82.70 &81.97 &79.00 &79.02 & 85.98 &85.98\\ \cline{2-8}						
				&PTE($G_{ww}+G_{wl}$)&83.90 &83.11 &81.65 &81.62 & 89.14 & 89.14\\ \cline{2-8}
				&PTE($G_{wd}+G_{wl}$)&\textbf{84.39} &\textbf{83.64} &82.29 & 82.27 & 89.76 & 89.76\\ \cline{2-8}
				&PTE(pretrain)&82.86 &82.12  &79.18 &79.21 & 86.28 & 86.28 \\ \cline{2-8}
				&PTE(joint)&84.20 &83.39 &\textbf{82.51} &\textbf{82.49} & \textbf{89.80} & \textbf{89.80}\\ \hline
			\end{tabular}
		}
	\end{center}
\end{table*}

\begin{table*} [!htdb]
	\caption{Results of text classification on \textbf{long} documents (RCV1 data sets). }
	\label{fig::results-long2}
	\begin{center}
		\scalebox{0.9}{
			\begin{tabular}{c|c|c|c|c|c|c|c|c}\hline
				& \multicolumn{2}{c|}{Corporate} & \multicolumn{2}{|c|}{Economics}& \multicolumn{2}{|c}{Government}& \multicolumn{2}{|c}{Market}\\\hline
				Algorithm&  Micro-F1& Macro-F1 &  Micro-F1& Macro-F1 &  Micro-F1& Macro-F1  &  Micro-F1& Macro-F1 \\ \hline \hline
				BOW&78.45 &63.80 &86.18 &81.67 &77.43 &62.38 &95.55 &94.09\\ \hline
				PVDBOW&65.87 &45.78 &79.63 &74.82 &70.74 &54.08 &91.81 &88.88\\ \hline			
				LINE($G_{wd}$)&76.76 &60.30 &85.55 & 81.46&77.82 &63.34 &95.66 &93.90\\ \hline
				PTE($G_{wl}$)&76.69 &60.48 &84.88 &80.02 &78.26 &63.69 &95.58 &93.84\\ \hline
				PTE(pretrain)&77.03 &61.03 &84.95& 80.63&78.48 &64.50 &95.54 &93.79\\ \hline										
				PTE(joint)&\textbf{79.20} &\textbf{64.29} &\textbf{87.05} & \textbf{83.01}&\textbf{79.63} &\textbf{66.15} &\textbf{96.19} &\textbf{94.58}\\ \hline			
			\end{tabular}
		}
	\end{center}
\end{table*}

\begin{table*} [!htdb]
	\caption{Results of text classification on \textbf{short} documents.}
	\label{fig::results-short}
	\begin{center}
		\scalebox{0.9}{
			\begin{tabular}{l|c|c|c|c|c|c|c}\hline
				&& \multicolumn{2}{c|}{DBLP} & \multicolumn{2}{|c|}{MR}& \multicolumn{2}{|c}{Twitter}\\\hline
				Type&	Algorithm&  Micro-F1& Macro-F1 &  Micro-F1& Macro-F1 &  Micro-F1& Macro-F1  \\ \hline \hline
				\scriptsize Word &		BOW&75.28 & 71.59&71.90 &71.90 &75.27 &75.27\\ \hline
				\multirow{6}{0.6in}{\scriptsize Unsupervised \\ Embedding}		&			Skip-gram&73.08 &68.92 &67.05 &67.05 &73.02 & 73.00\\ \cline{2-8}							
				&PVDBOW&67.19 & 62.46&67.78 &67.78 &71.29 &71.18 \\ \cline{2-8}			
				&PVDM&37.11 &34.38 &58.22 &58.17 &70.75 &70.73 \\ \cline{2-8}						
				&LINE($G_{ww}$)&73.98 &69.92 &71.07 &71.06 & 73.19&73.18 \\ \cline{2-8}
				&LINE($G_{wd}$)&71.50 &67.23 &69.25 &69.24 &73.19 &73.19\\ \cline{2-8}
				&LINE($G_{ww}+G_{wd}$)&74.22 &70.12 &71.13 &71.12 & 73.84&73.84 \\ \hline
				\multirow{7}{0.6in}{\scriptsize Predictive \\ Embedding}			&CNN&76.16 &73.08  &72.71 &72.69 &\textbf{75.97} &\textbf{75.96} \\ \cline{2-8}
				&CNN(pretrain)&75.39 &72.28&68.96 & 68.87   &75.92 & 75.92  \\ \cline{2-8}					
				&PTE($G_{wl}$)&76.45 &72.74 &73.44 &73.42 & 73.92&73.91 \\ \cline{2-8}	
				&PTE($G_{ww}+G_{wl}$)&76.80 &73.28 &72.93 &72.92 &74.93 & 74.92\\ \cline{2-8}
				&PTE($G_{wd}+G_{wl}$)&\textbf{77.46} &\textbf{74.03}& 73.13&73.11 &75.61 &75.61\\ \cline{2-8}
				&PTE(pretrain)&76.53 &72.94 & 73.27&73.24  &73.79 &73.79\\ \cline{2-8}			
				&PTE(joint)&77.15 &73.61 & \textbf{73.58}&\textbf{73.57} &75.21 &75.21\\ \hline
			\end{tabular}
		}
	\end{center}
\end{table*}

\begin{figure*}[htdb!]
	\centering
	\subfigure[\textsc{20ng}]{
		\label{fig::performance_vs_label_20ng}
		\includegraphics[width=0.22\textwidth]{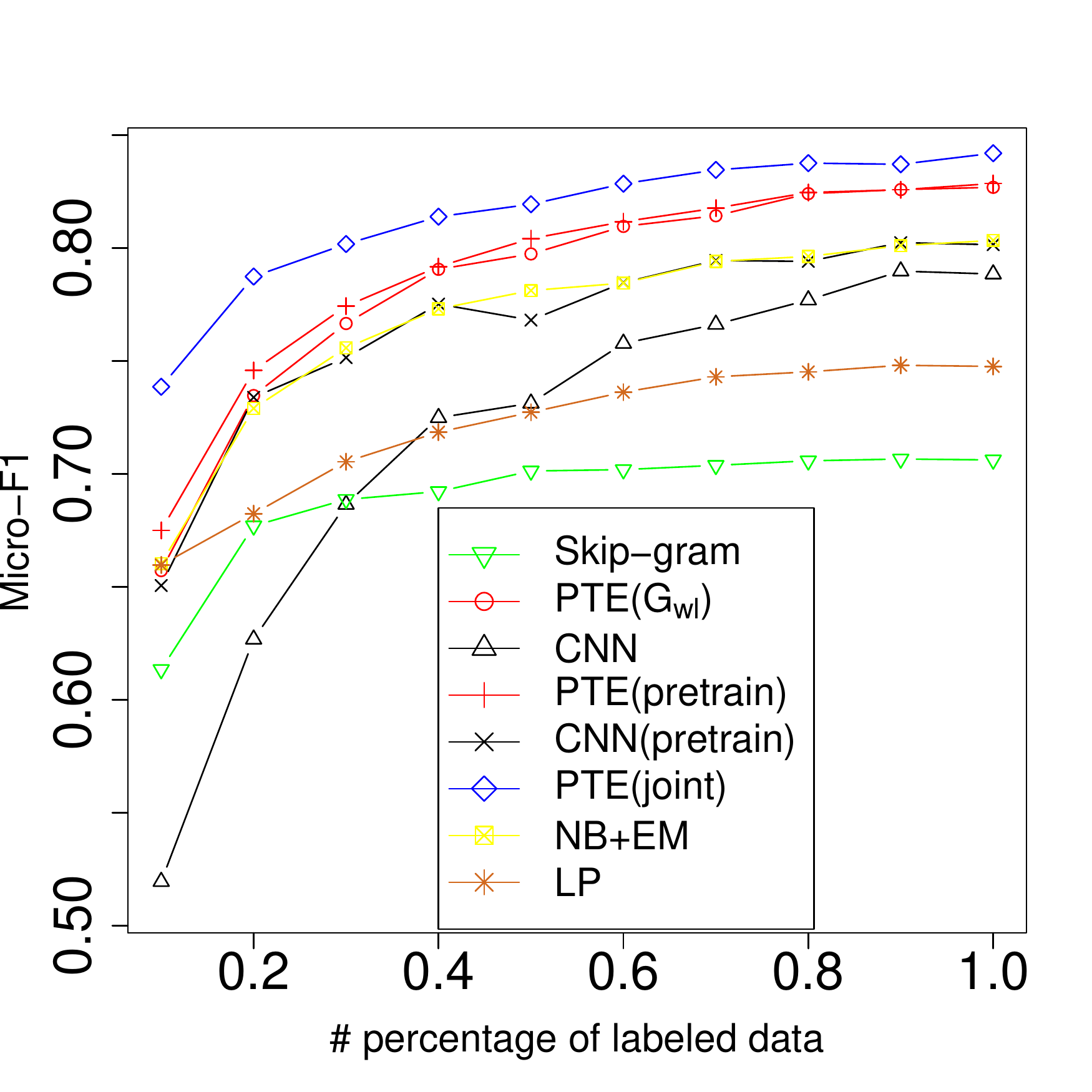}
	}
	\subfigure[\textsc{imdb}]{
		\label{fig::performance_vs_label_imdb}
		\includegraphics[width=0.22\textwidth]{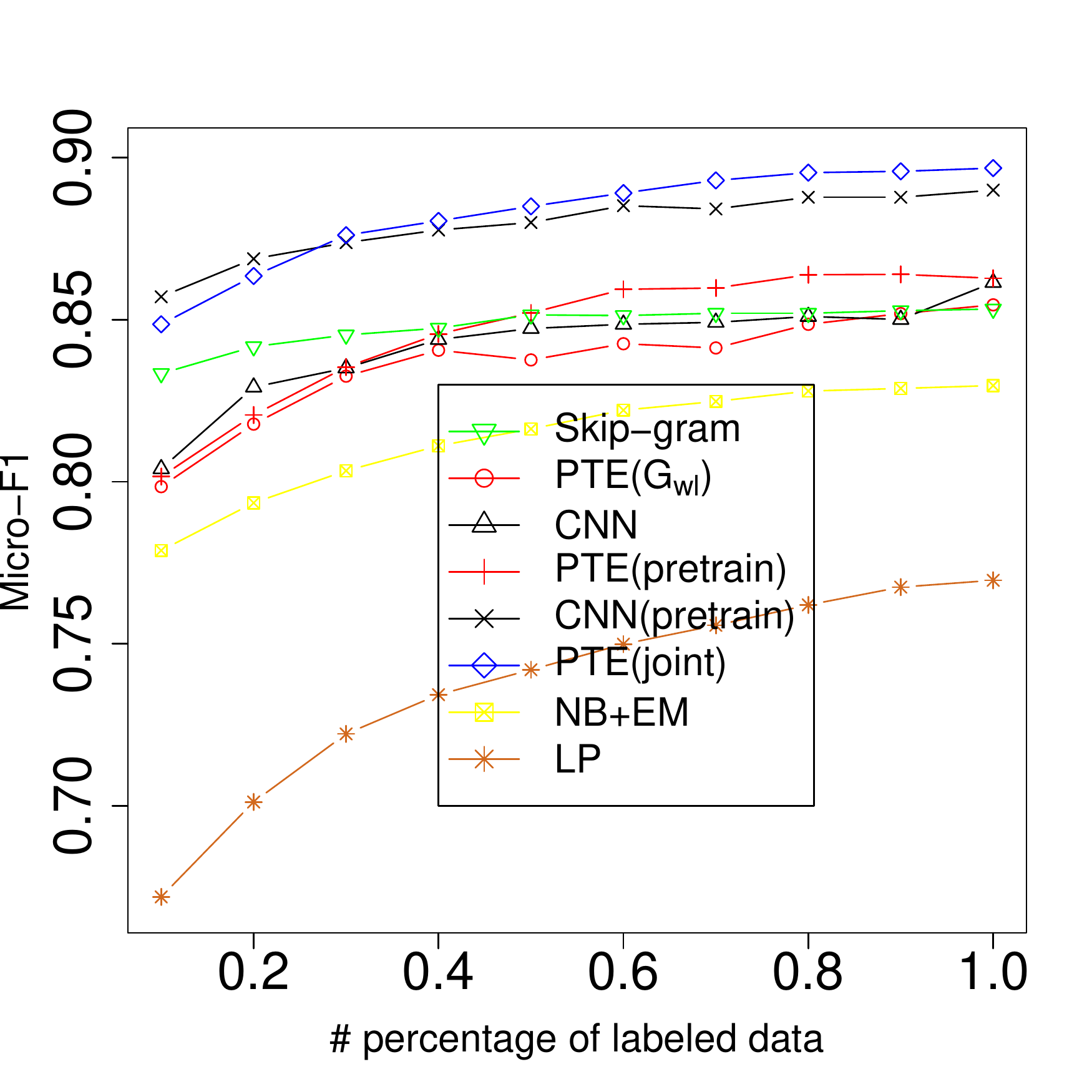}
	}
	\subfigure[\textsc{dblp}]{
		\label{fig::performance_vs_label_dblp}
		\includegraphics[width=0.22\textwidth]{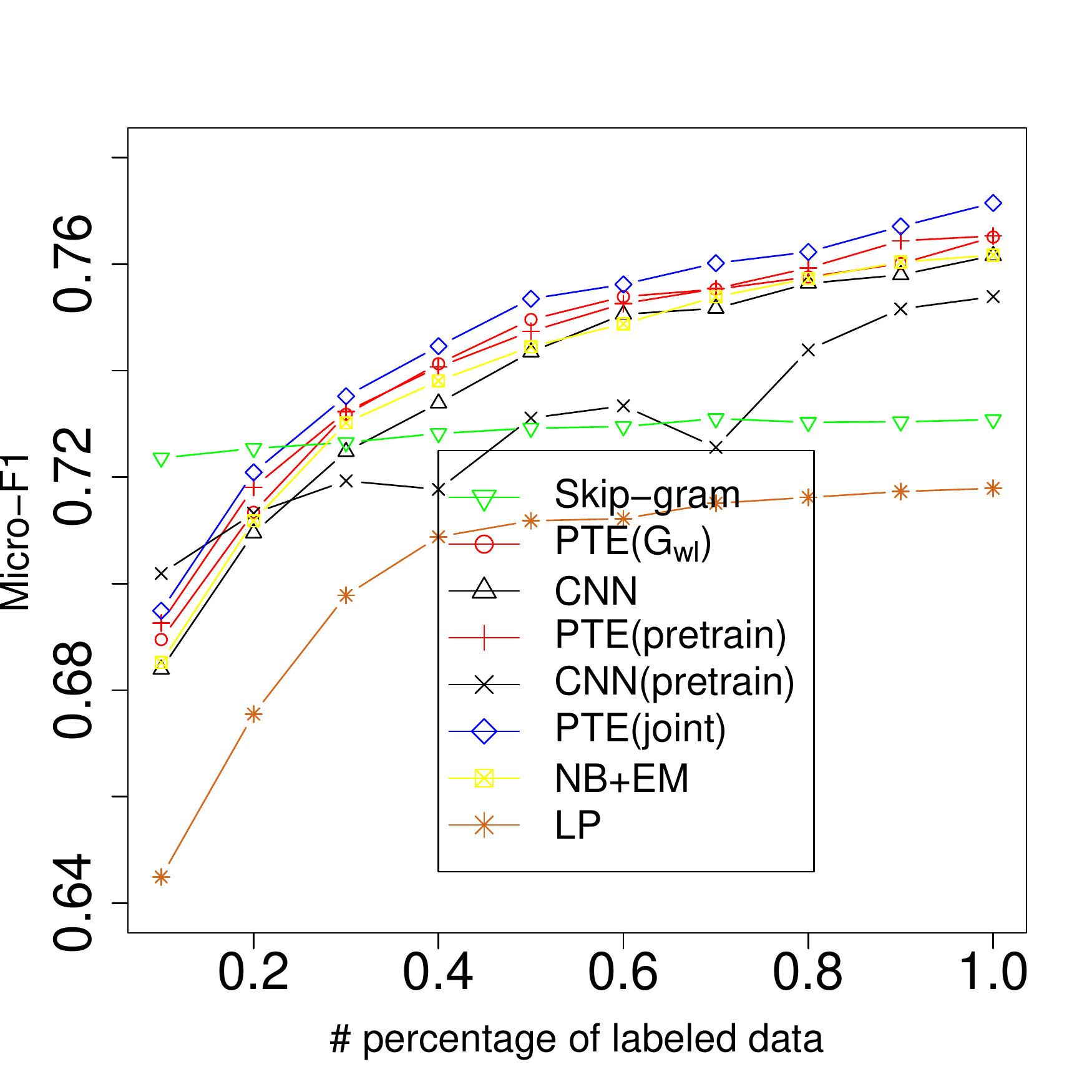}
	}
	\subfigure[\textsc{mr}]{
		\label{fig::performance_vs_label_dblp}
		\includegraphics[width=0.22\textwidth]{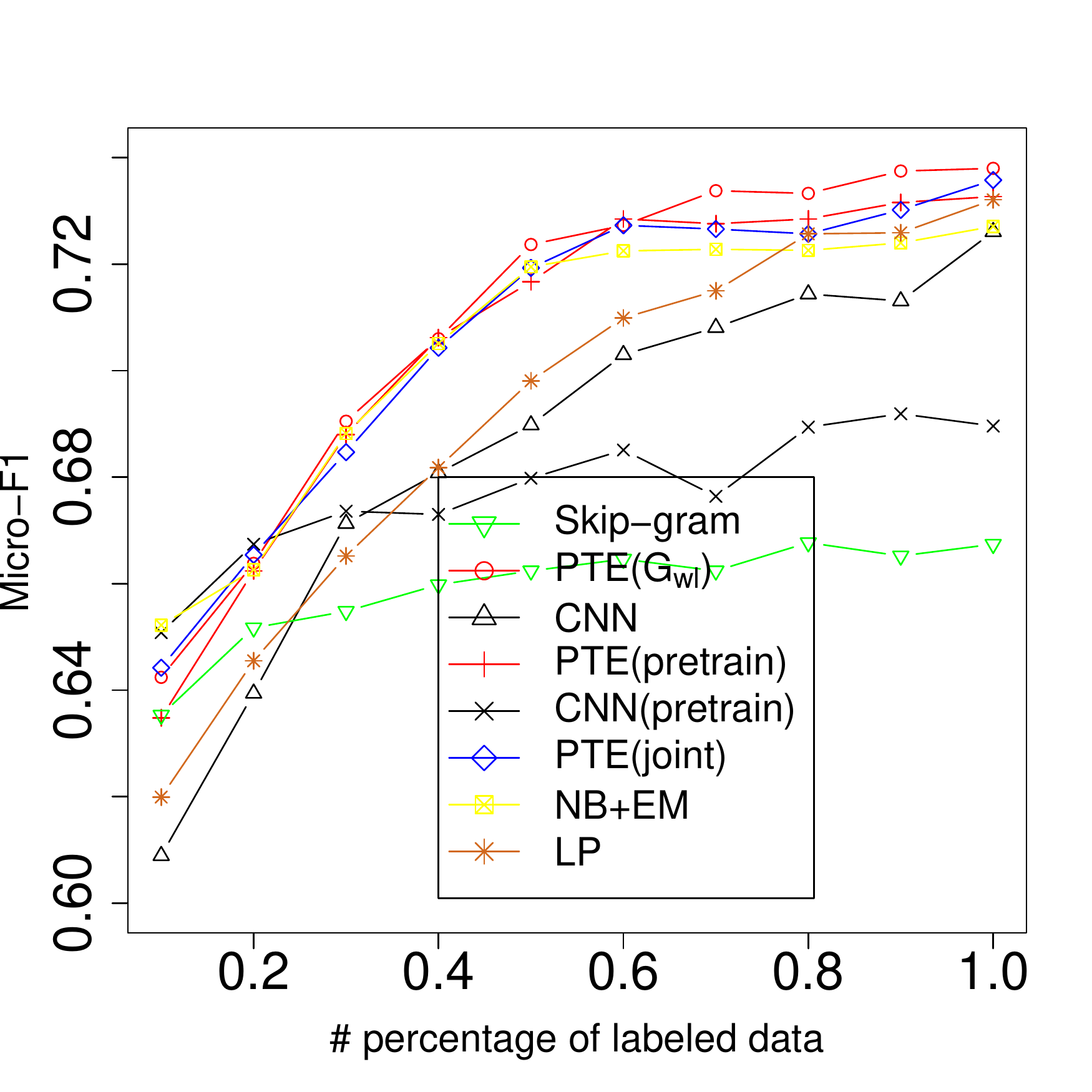}
	}			
	\caption{Performance w.r.t. \# labeled data.}
	\label{fig::results_labeled}
\end{figure*}

Table~\ref{fig::results-long1} and~\ref{fig::results-long2} compare the performance of text classification on long documents. Let us start with Table~\ref{fig::results-long1} on \textsc{20ng}, \textsc{Wiki} and \textsc{Imdb} data sets. We first compare the performance of unsupervised embedding methods, which use either local word co-occurrences (Skip-gram, LINE($G_{ww}$)), document level word co-occurrences (PV-DBOW, LINE($G_{wd}$)), or the combination (LINE($G_{ww}+G_{wd}$)). We can see that the LINE($G_{wd}$) with document-level word co-occurrences performs the best among the unsupervised embeddings. 
The performance of PVDM is inferior to that of PVDBOW, which is different from what is reported in~\cite{le2014distributed}. Unfortunately we are not able to replicate their results. Similar results to ours are also reported in~\cite{liu2015topical}. For the results of PVDBOW on the \textsc{imdb} data set, our results are different from those reported in~\cite{le2014distributed,mesnil2014ensemble}. This is because their embeddings are trained on the mixture of training and test data sets while our embeddings are only trained with the training data, which we believe is a more reasonable experiment setup.  

Next we compare the performance of predictive embeddings, including different variants of CNN and PTE. When PTE is jointly trained using the heterogeneous text network or with the combination of word-document and word-label networks, it performs the best among all the approaches. All PTE approaches jointly trained with the word-label network (e.g., $G_{ww}+G_{wl}$) outperform their corresponding unsupervised embedding approaches (e.g., $G_{ww}$), which shows the power of learning predictive text embeddings with the supervision. PTE(joint) consistently outperforms PTE($G_{wl}$), demonstrating that incorporating unlabeled information, i.e., word-word and word-document networks, also improves the quality of the embeddings. PTE(joint) also significantly outperforms PTE(pretrain). This shows that jointly training with unlabeled and labeled data is much more effective compared to separating them into two phases of pre-training and fine-tuning.

It is interesting to observe that PTE(joint) consistently outperforms CNN. This is promising as PTE does not use a sophisticated neural network architecture. We also attempt to pre-train the CNN with the word embeddings learned by the LINE($G_{wl}+G_{wd}$) and then fine tune it with the labels. Surprisingly, the performance of CNN with pre-training significantly improves on the \textsc{20ng} and \textsc{imdb} data sets and remains almost the same on the \textsc{wiki} data set. This implies that pre-training CNN with a well learned unsupervised embeddings can be very useful. However, even with pre-training the performance of CNN is still inferior to that of PTE(joint). This is probably because the PTE model can jointly train with both the unlabeled and labeled data while CNN can only utilize them separately through pre-training and fine-tuning. 
PTE(joint) also outperforms the classical ``bag-of-words'' representation even though the dimensionality of the embeddings learned by the PTE is way smaller than that of ``bag-of-words.''

Table~\ref{fig::results-long2} reports the results on the RCV1 data sets. As the order between the words is lost, the embedding methods that require the word order information are not applicable. Similar results are observed.  Predictive text embeddings outperform unsupervised embeddings. PTE(joint) is also much more effective than PTE(pretrain).

All the embedding approaches (except ``bag-of-words'') are trained with asynchronous stochastic gradient descent algorithm using 20 threads on a single machine with 1T memory, 40 CPU cores at 2.0GHZ. We compare the running time of CNN and PTE(joint) on the \textsc{imdb} data set. The PTE(joint) method is typically more than 10 times faster than the CNN models. When pre-trained with preexisting word embeddings, CNN converges much faster, but still more than 5 times slower than PTE(joint). 


\subsubsection{Performance on Short Documents}

Table~\ref{fig::results-short} compares the performance on short documents. Among unsupervised embeddings, the LINE$(G_{ww}+G_{wd})$, which combines the document-level and local context-level word co-occurrences, performs the best. The LINE($G_{ww}$) utilizing the local context-level word co-occurrences outperforms the LINE($G_{wd}$) using document-level word co-occurrences, which is opposite to the observations on long documents. This is because document-level word co-occurrences suffer from the sparsity in short documents, with similar results observed in statistical topic models~\cite{tang2014understanding}. The performance of PVDM is still inferior to that of PVDBOW, which is consistent with the results on long documents.


For predictive embeddings, the best performance is obtained by the PTE (on \textsc{dblp}, \textsc{mr} ) or CNN (on \textsc{twitter}). Among the PTE approaches, the predictive embeddings learned by incorporating the word-label network outperform the corresponding unsupervised embeddings, which is consistent with the results on long documents. PTE(joint) outperforms LINE($G_{wl}$), showing the usefulness of incorporating unlabeled information. PTE(joint) also significantly outperforms the PTE(pretrain), showing the advantage of jointly training with the labeled and unlabeled data. 

On the short documents, we observe that PTE(joint) does not consistently outperform the CNN. The reason is probably due to the problem of word sense ambiguity, which becomes more serious on the short documents. CNN reduces the problem of word ambiguity through using the word orders in local context in the convolutional kernels while PTE does not leverage the orders. We believe there is considerable room to improve predictive text embedding by utilizing word orders, which we leave as future work.

\subsection{Effects of Labeled Data}

We compare CNN and PTE head-to-head by varying the percentages of labeled data. We consider the cases without or with unlabeled data, mimicking the scenarios of supervised and semi-supervised learning.  In the setting of semi-supervised learning, we also compare with classical semi-supervised approaches, Naive Bayes with EM (NB+EM)~\cite{su2011large} and label propagation (LP)~\cite{zhou2004learning}.  Fig.~\ref{fig::results_labeled} reports the performance on both long and short documents. Overall, both CNNs and PTEs improve when the size of labeled data increases. In the supervised settings, i.e., between CNN and PTE($G_{wl}$), PTE($G_{wl}$) outperforms or is comparable to CNN on both the long and short documents. In the semi-supervised settings, i.e., between CNN(pretrain) and PTE(joint), PTE(joint) consistently outperforms CNN(pretrain), which is pre-trained with the \textbf{best performing} unsupervised word embeddings. The PTE(joint) also outperforms the state-of-the-art semi-supervised approaches Naive Bayes with EM and label propagation.

We also notice that when the size of labeled data is scarce, pre-training CNN with unsupervised embeddings is quite helpful, especially on the short documents. It even outperforms all PTEs when the training examples are too few. However, when the size of labeled data increases, pre-training CNN does not always improve its performance (e.g., on the \textsc{dblp} and \textsc{mr} data sets). 

Note that for Skip-gram, increasing the number of labeled data in training does not further increase the performance. We also notice that when the labeled documents are too few, the performance of PTE is inferior to the Skip-gram on the \textsc{dblp} data set. The reason is that when the number of labeled examples is scarce, the word-label network is very noisy and PTE treats the word-label network equally to the robust word-word/word-document networks. A way to fix is to adjust the sampling probability from the word-label and word-word/word-document when the labeled data is scarce. We leave it as future work.

\subsection{Effects of Unlabeled Data}

\vskip -1.5em
\begin{figure}[htdb!]
	\centering
	\subfigure[\textsc{20ng}]{
		\label{fig::performance_vs_unlabel_20ng}
		\includegraphics[width=0.22\textwidth]{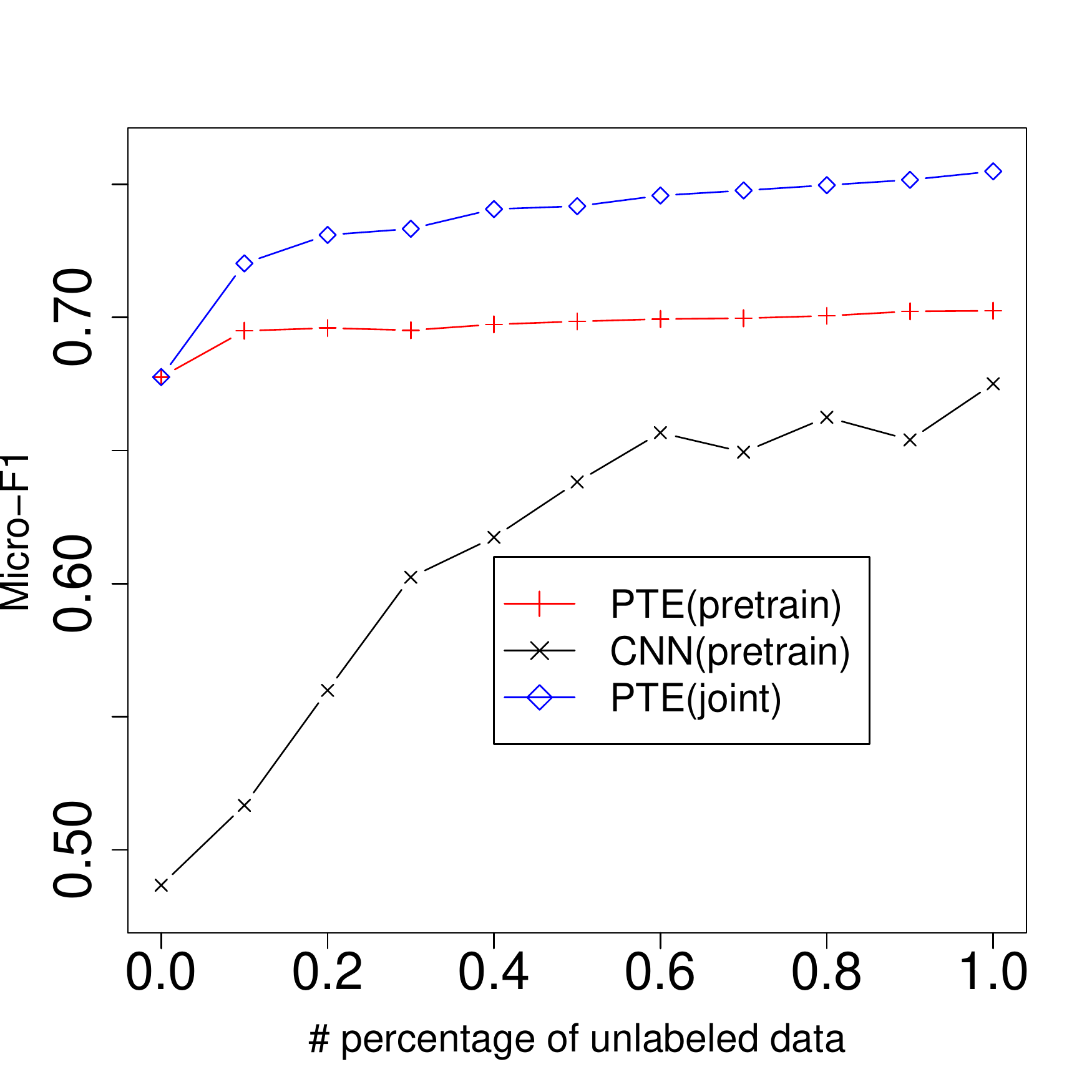}
	}
	\subfigure[\textsc{dblp}]{
		\label{fig::performance_vs_unlabel_dblp}
		\includegraphics[width=0.22\textwidth]{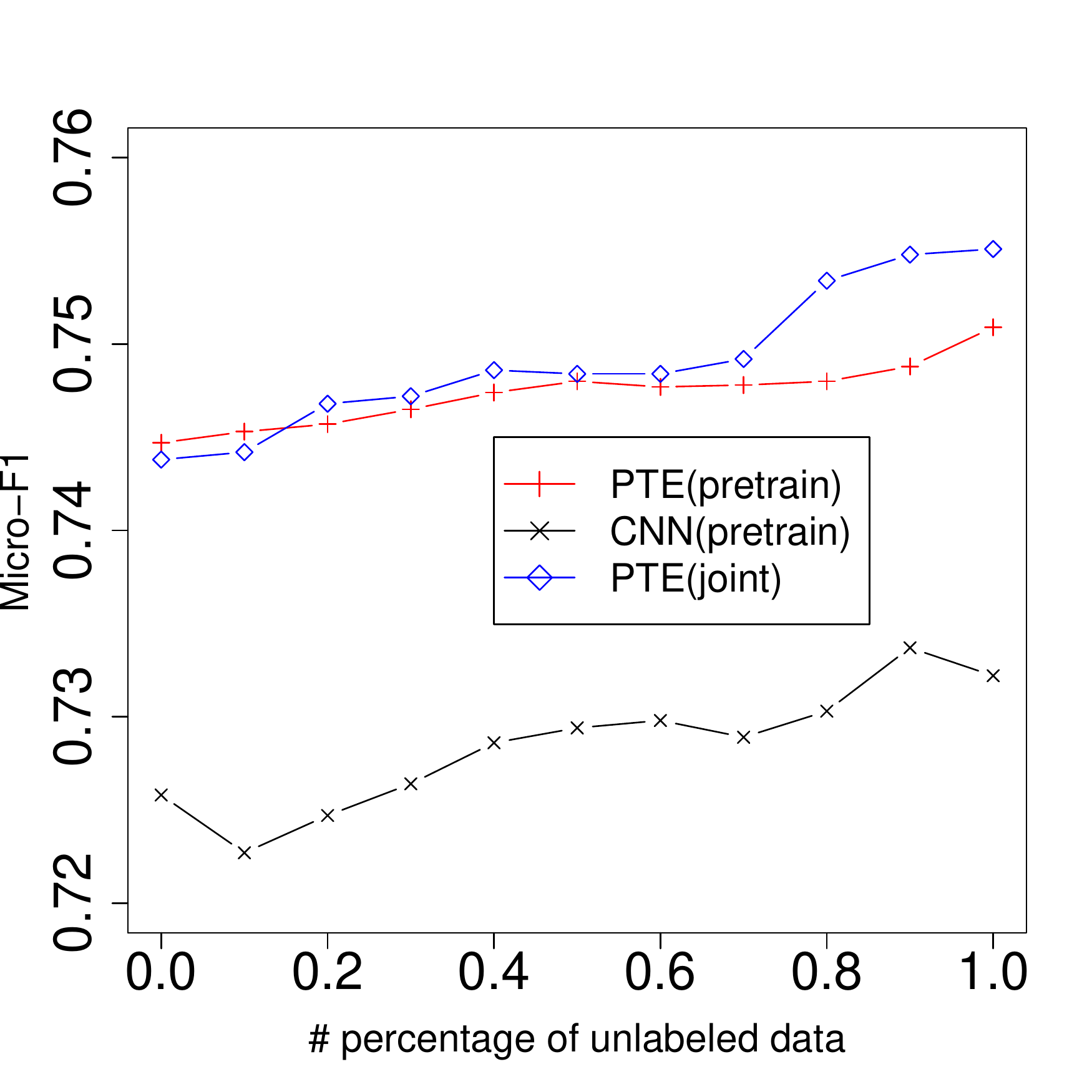}
	}
	\caption{Performance w.r.t. \# unlabeled data.}
	\label{fig::results_unlabeled}
\end{figure}

We also analyze the performance of the CNNs and PTEs w.r.t. the size of unlabeled data. For CNN, the unlabeled data is used for pre-training while for the PTE, the unlabeled data can be used for either pre-training or jointly training. Fig.~\ref{fig::results_unlabeled} reports the results on \textsc{20ng} and \textsc{dblp} data sets. Due to space limitation, we omit the results on other data sets, which are similar. On \textsc{20ng}, we use 10\% documents as labeled while the rest is used as the unlabeled; on \textsc{dblp}, we randomly sample 200,000 titles of the papers published in the other conferences as the unlabeled data. We use the unsupervised embeddings (learned by LINE$(G_{ww}+G_{wd})$) as the pre-training of CNN. We can see that the performance of both CNN and PTE improves when the size of unlabeled data increases. For PTE, the way of jointly training with the unlabeled and labeled data is much more effective than separating them into pre-training and fine-tuning.


\subsection{Parameter Sensitivity}
\vskip -1.5em
\begin{figure}[htdb!]
	\centering
	\subfigure[\textsc{20ng}]{
		\label{fig::sensitivity_20ng}
		\includegraphics[width=0.22\textwidth]{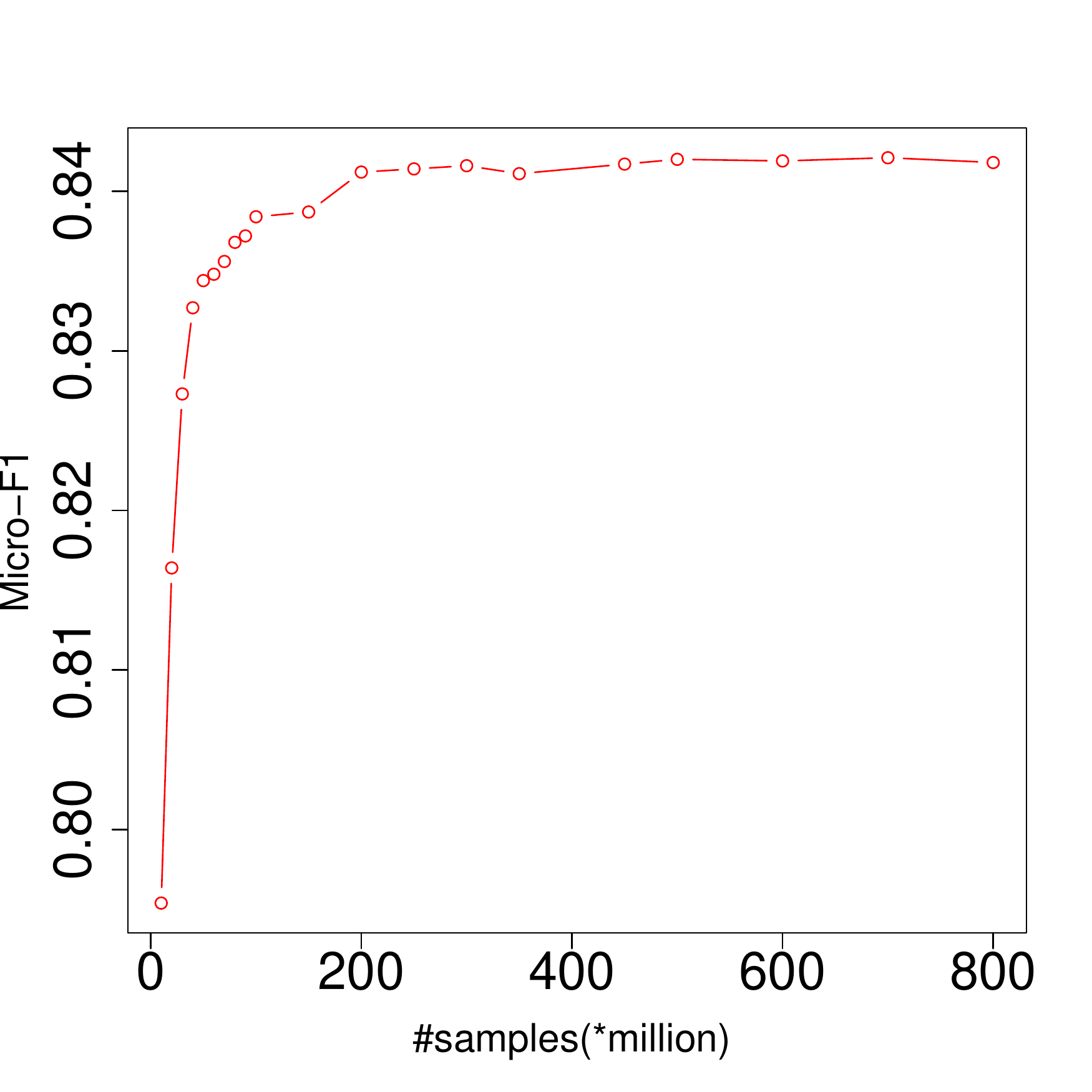}
	}
	\subfigure[\textsc{dblp}]{
		\label{fig::sensitivity_dblp}
		\includegraphics[width=0.22\textwidth]{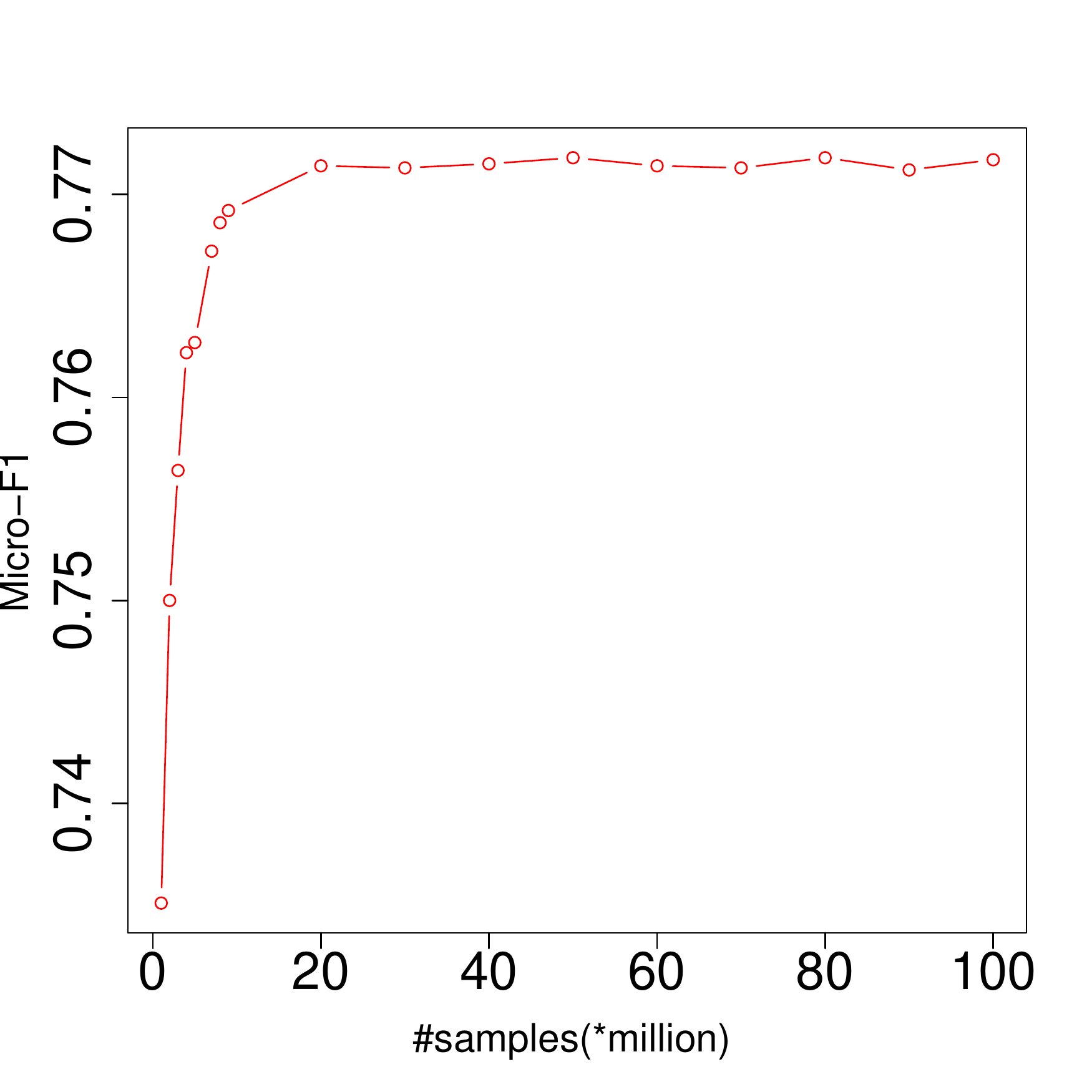}
	}
	\caption{Sensitivity of PTE w.r.t. number of samples $T$.}
	\label{fig::sensitivity}
\end{figure}

For the proposed PTE models, as mentioned previously, most of the parameters except for the number of edge samples $T$ are not sensitive to different data sets and can be set by default. Here we analyze the performance sensitivity of PTE(joint) w.r.t the number of samples $T$. Fig.~\ref{fig::sensitivity} reports the results on the \textsc{20ng} and \textsc{dblp} data sets. On both data sets, we can see that when the number of samples $T$ becomes large enough, the performance of PTE(joint) converges. Therefore, in practice, one can just set the number of samples $T$ to be sufficiently large.  A reasonable estimation of $T$ we find in practice is several times of the number of edges in the heterogeneous text networks.

\begin{figure}[htdb!]
	\centering
	\subfigure[Train(LINE($G_{wd}$))]{
		\label{fig::visualization-train-dw}
		\includegraphics[width=0.22\textwidth]{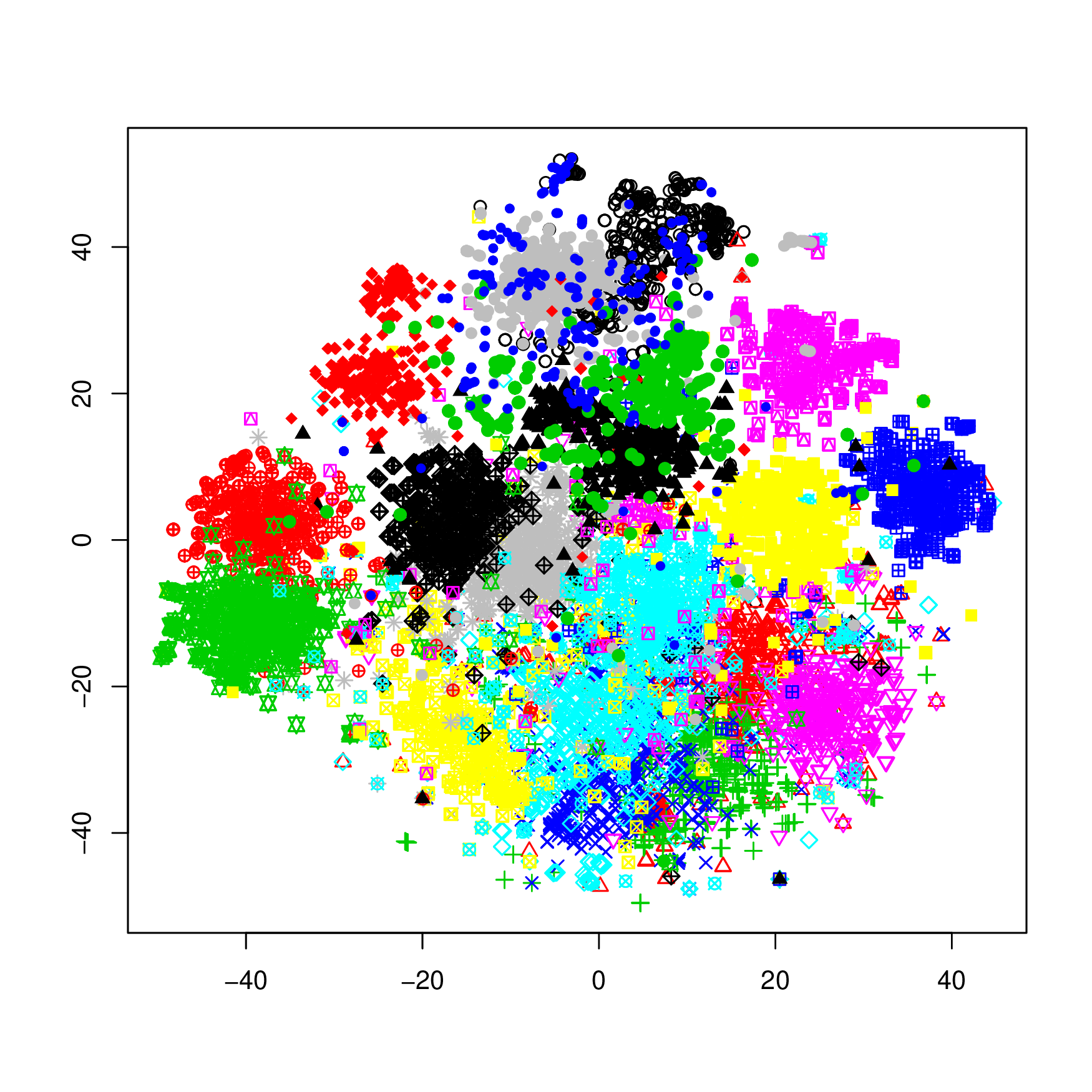}
	}
	\subfigure[Train(PTE($G_{wl}$))]{
		\label{fig::visualization-train-lw}
		\includegraphics[width=0.22\textwidth]{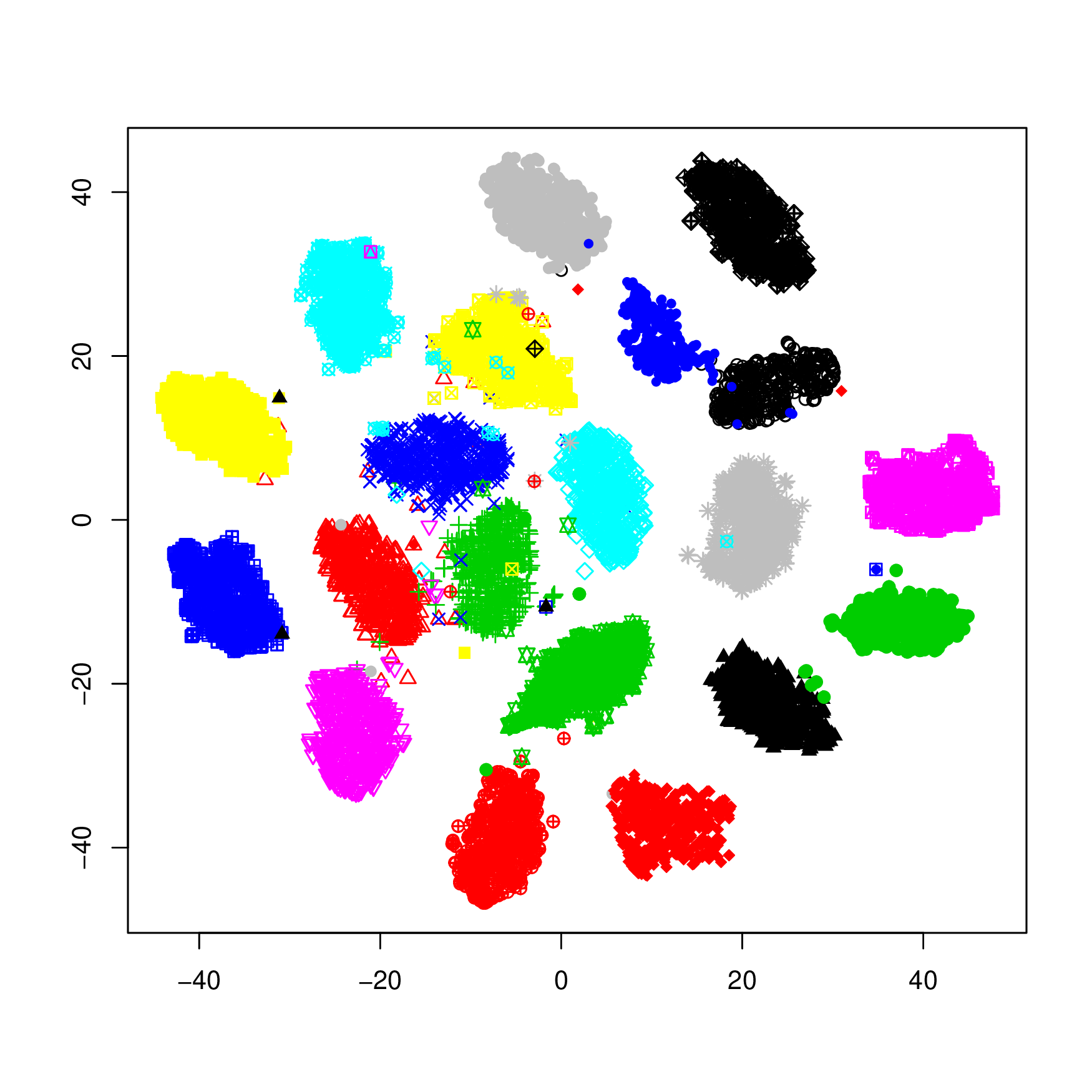}
	} 
	\subfigure[Test(LINE($G_{wd}$))]{
		\label{fig::visualization-test-dw}
		\includegraphics[width=0.22\textwidth]{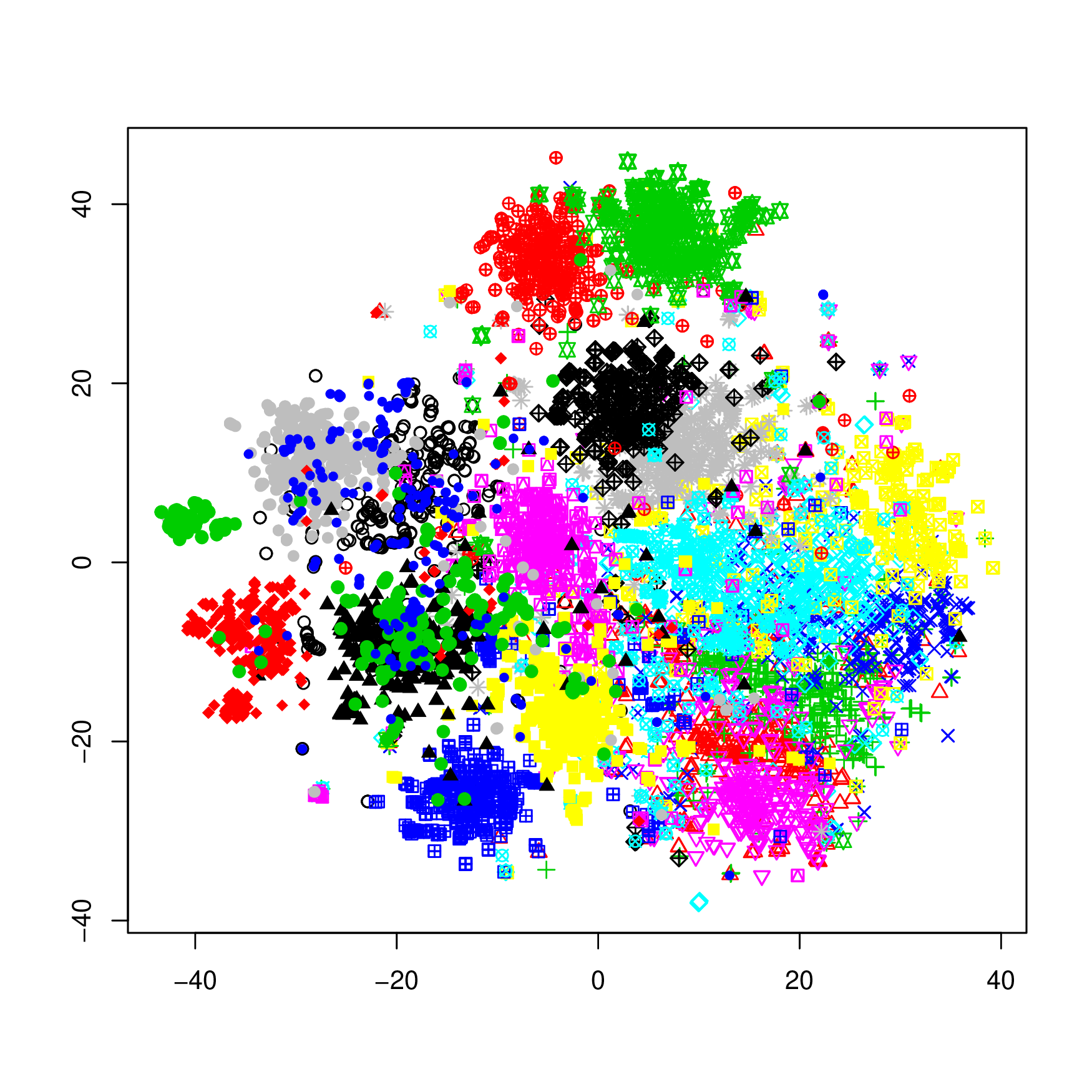}
	}
	\subfigure[Test(PTE($G_{wl}$))]{
		\label{fig::visualization-test-lw}
		\includegraphics[width=0.22\textwidth]{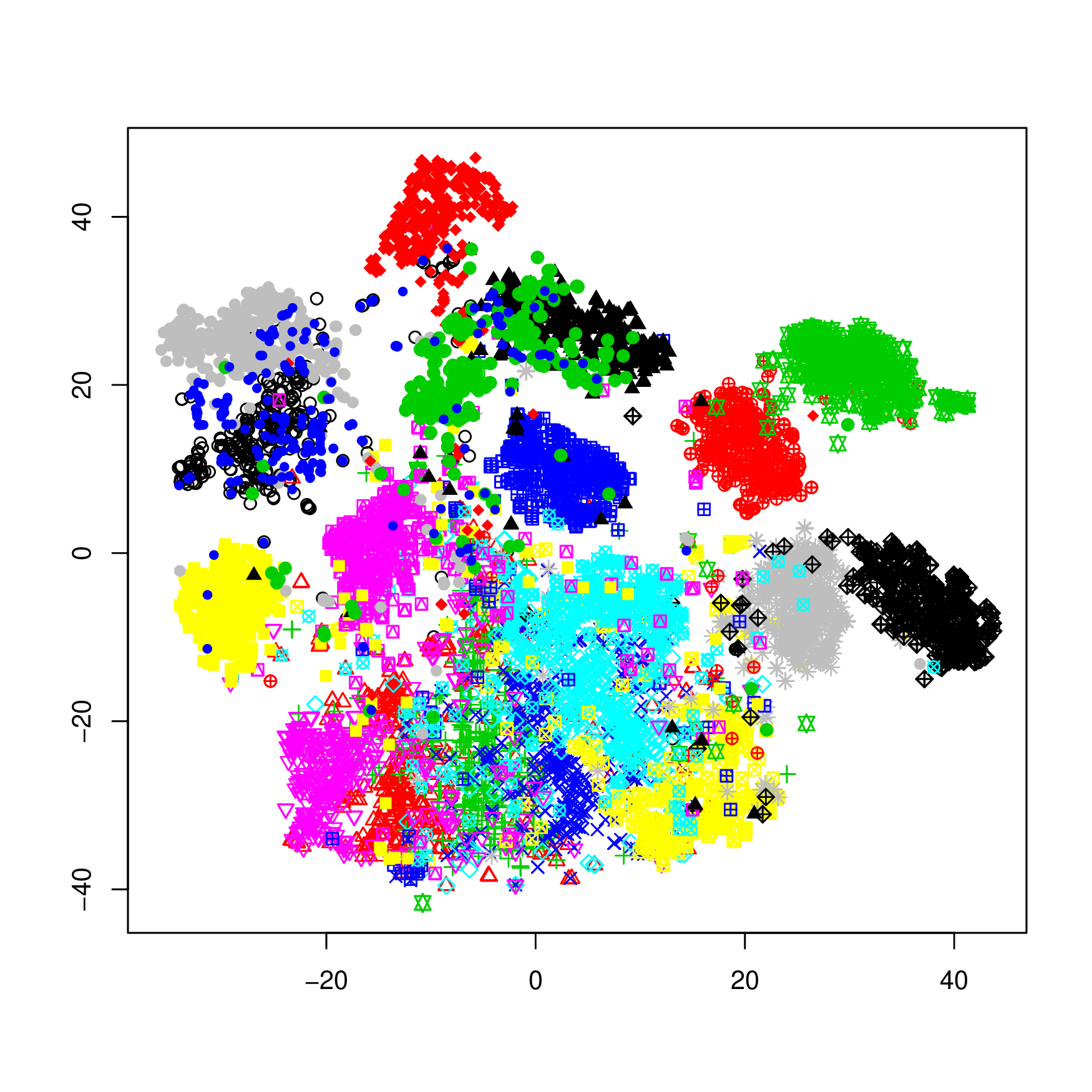}
	} 
	\caption{Document visualization using unsupervised and predictive embeddings on \textsc{20ng} data set, visualized with the t-SNE tool~\cite{van2008visualizing}.}
	\label{fig::visualization-documents-20ng}	
\end{figure}

\subsection{Document Visualization}
Finally, we give an illustrative visualization of the documents in \textsc{20ng} to compare the unsupervised and predictive embeddings. We used LINE$(G_{wd})$ for the unsupervised embedding and PTE$(G_{wl})$ for the predictive embedding. Both the training and test documents are visualized. Fig.~\ref{fig::visualization-documents-20ng} shows the visualization results. We can see that on both training and test data sets, the predictive embedding much better distinguishes different classes than those learned by the unsupervised embedding, which intuitively shows the power of predictive embeddings on the task of text classification.

\section{Discussion and Conclusion}
\label{sec::conclusion}

\noindent \textbf{Unsupervised embeddings.} The essential information used by the unsupervised approaches is the local context-level or document-level word co-occurrences. On long documents, we can see that document-level word co-occurrences are more useful than the local context-level word co-occurrences, and the combination of two does not further improve the result; on short documents, the local context-level word co-occurrences are more useful than the document-level word co-occurrences, and their combination will further improve the embedding. Document-level word co-occurrences suffers from short lengths of documents. 

\noindent \textbf{Predictive embeddings.} Comparing between CNN and PTE, the CNN model seems to handle labeled information more effectively, especially on short documents. This is because CNN uses a much more complicated structure than the PTE, which in particular utilizes word orders in the local context and addresses word sense ambiguity.

Therefore, in the case when labeled data is very sparse, CNN can outperform the PTE, especially on short documents. However, this advantage is at the expense of intensive computation and exhaustive parameter tuning. On the other hand PTE is much faster and much easier to configure (with few parameters to tune). When the labeled data becomes abundant, the performance of PTE will be at least comparable and usually superior to CNN. 

Compared to the CNN model, an obvious advantage of the PTE model is its capability of being able to jointly train with both the unlabeled and labeled data. CNN can only make use of unlabeled data through an indirect way, i.e., pre-training with the unsupervised word embeddings learned from other methods. Pre-training may not always help when the size of labeled data becomes abundant. 

\paragraph{\noindent \textbf{Practical Guidelines}} Based on the above discussions,  we provide the following practical guidelines, which suggest to choose between CNN or PTE in different scenarios.

\begin{description}
	\item [(1)] When no labeled data is available, we suggest using LINE$(G_{wd})$ for learning an unsupervised word embedding from long documents and using LINE($G_{wd}+G_{ww}$) from short documents.
	\item [(2)]When few labeled data is available, on short documents, we suggest to learn an unsupervised embedding first (according to the first guideline) and then pre-train CNN with the unsupervised word embedding; on long documents, we suggest using PTE. 
	\item [(3)]When the labeled data are abundant, on long documents, we strongly suggest using the PTE model to jointly train the embedding with both the unlabeled and labeled data; on short documents, the selection between PTE(joint) and CNN or CNN(pretrain) basically trades performance with efficiency. 
\end{description}

We believe this study provides an interesting direction to efficiently learn predictive, distributed text embeddings. It is feasible and desirable to develop similar methods that compete with deep neural networks end-to-end on specific tasks but avoid complex model architectures or intensive computation. There is considerable room to improve PTE, for example by considering the orders of words.

\section*{Acknowledgments}
Qiaozhu Mei is supported by the National Science Foundation under grant numbers IIS-1054199 and CCF-1048168. 

\bibliographystyle{abbrv}

\end{document}